\documentclass{article}

\usepackage[preprint]{neurips_2025}
\PassOptionsToPackage{numbers, compress}{natbib}

\usepackage[utf8]{inputenc} 
\usepackage[T1]{fontenc}    
\usepackage{hyperref}       
\usepackage{url}            
\usepackage{booktabs}       
\usepackage{amsfonts}       
\usepackage{nicefrac}       
\usepackage{microtype}      
\usepackage{xcolor}         
\usepackage{amsmath}
\usepackage{multirow}
\usepackage{graphicx}
\usepackage{tcolorbox} 
\tcbuselibrary{breakable}

\newcommand{\method}[0]{\textsc{CC-Learn}}

\title{\method{}: Cohort-based Consistency Learning}

\author{
  Xiao Ye ~~~~ Shaswat Shrivastava ~~~~ Zhaonan Li ~~~~ Jacob Dineen \\
   \textbf{~~~~ Shijie Lu ~~~~ Avneet Ahuja ~~~~ Ming Shen ~~~~ Zhikun Xu ~~~~ Ben Zhou}\\
  Arizona State University\\
  \texttt{xiaoye2@asu.edu}
}

\begin{document}

\maketitle

\begin{abstract}
Large language models excel at many tasks but still struggle with consistent, robust reasoning. We introduce Cohort-based Consistency Learning (\method{}), a reinforcement learning framework that improves the reliability of LLM reasoning by training on \emph{cohorts of similar questions} derived from shared programmatic abstractions. To enforce cohort-level consistency, we define a composite objective-combining cohort accuracy, a retrieval bonus for effective problem decomposition, and a rejection penalty for trivial or invalid lookups—that reinforcement learning can directly optimize, unlike supervised fine-tuning. Optimizing this reward guides the model to adopt uniform reasoning patterns across all cohort members. Experiments on challenging reasoning benchmarks (including ARC-Challenge and StrategyQA) show that \method{} boosts both accuracy and reasoning stability over pretrained and SFT baselines. These results demonstrate that cohort-level RL effectively enhances reasoning consistency in LLMs.
\end{abstract}

\begin{figure}[!htbp]
  \centering
  \includegraphics[width=\linewidth]{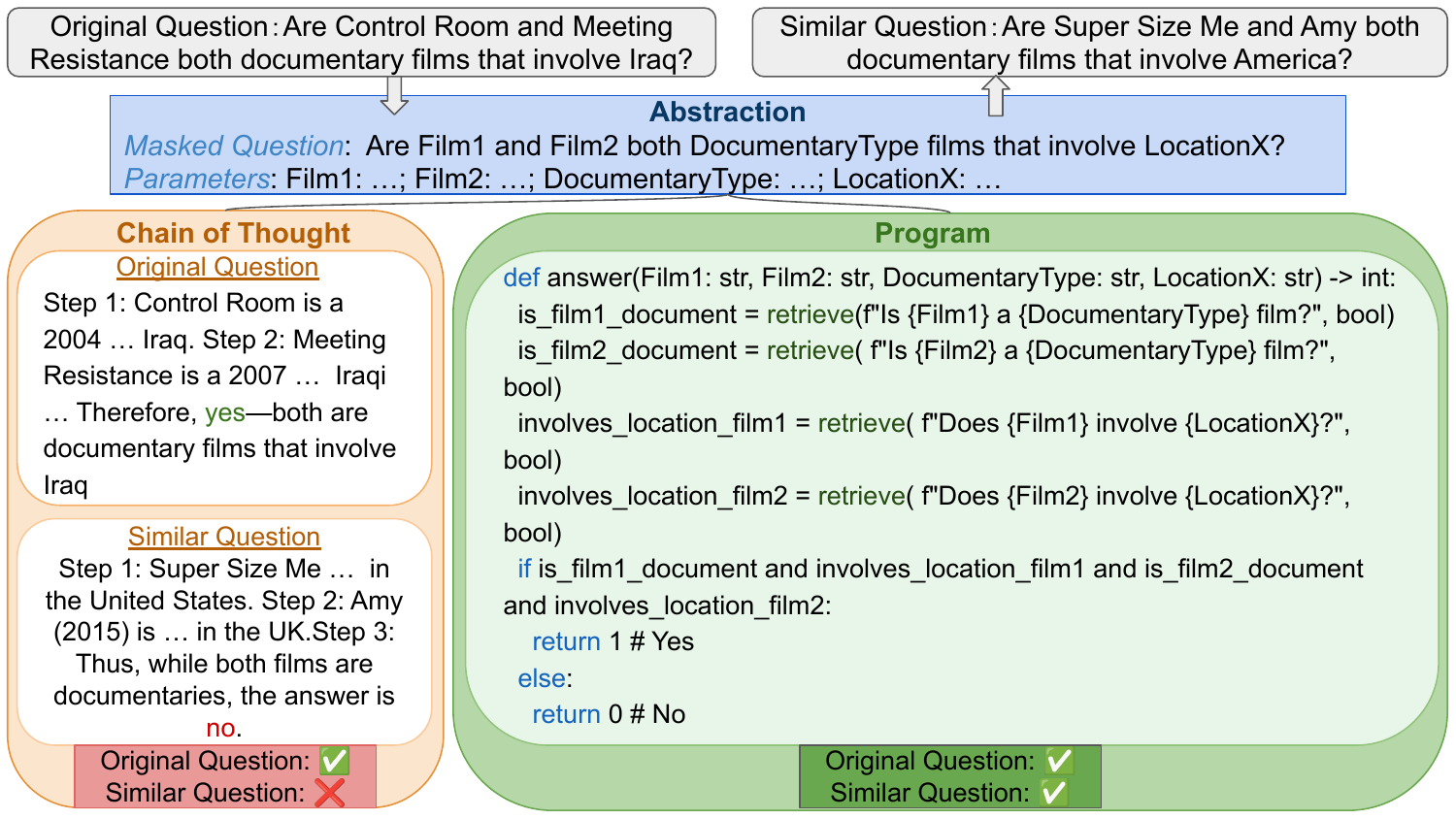}
  \caption{An illustrative example of cohort‐based consistency learning(See Appendix ~\ref{sec:screenshot}). (Top) The original question and a surface‐variant question sharing the same reasoning path. (Middle) The masked abstraction template with its parameter dictionary, used to generate a cohort of factual variants. (Bottom) The executable program synthesized by the model, which issues simple \texttt{retrieve} calls for each substep and computes the final answer, enforcing consistent reasoning across all cohort members.}
  \label{fig:example}
\end{figure}

\section[Introduction]{Introduction\protect\footnote{Code and data will be released with the camera‐ready version.}}

Large language models (LLMs) have made remarkable progress in complex reasoning tasks through strategies like prompting and step-by-step solution traces. Techniques such as \emph{chain-of-thought} prompting \citep{wei2022cot} enable models to decompose problems into intermediate steps, significantly improving performance on arithmetic, commonsense, and various reasoning challenges. Similarly, decoding strategies like self-consistency \citep{wang2023selfconsistency} enhance accuracy by sampling multiple reasoning paths and selecting the most consistent answer across benchmarks. Despite these advances, LLMs frequently exhibit \textbf{inconsistency}: a model may correctly answer a question in one formulation but fail on a paraphrase or logically equivalent variant\citep{yu2024reasonagainusingextractablesymbolic, zhou2024conceptualunbiasedreasoninglanguage, li2024deceptivesemanticshortcutsreasoning}. This behavior suggests brittle reasoning processes and undermines reliability in practical applications \citep{mccoy2019right, geirhos2020shortcut}.

Figure~\ref{fig:example} illustrates this phenomenon with an example from HotpotQA\citep{yang2018hotpotqa}.  The original question asks:
\begin{quote}
  “Are \textit{Control Room} and \textit{Meeting Resistance} both documentary films that involve Iraq?”  
\end{quote}
An LLM can follow a correct chain of thought for the original—but when presented with a similar variant,
\begin{quote}
  “Are \textit{Super Size Me} and \textit{Amy} both documentary films that involve America?”  
\end{quote}
it answers incorrectly by relying on an association between “Amy” and “UK” rather than applying the same logical checks \citep{sinha2021surface,zhao2021calibrate}. This gap between high single-question accuracy and poor generalization across variants highlights the need for training methods that enforce consistency.

To address this gap, we propose \method{}, which trains LLMs on \emph{cohorts} of similar questions expressed as executable programs, and optimizes a \emph{cohort-level} reward via reinforcement learning.  First, each question is transformed into a masked template exposing its core reasoning structure (e.g.,  
“Are \texttt{Film1} and \texttt{Film2} both \texttt{DocumentaryType} films that involve \texttt{Country}?”).  From this template we automatically generate a cohort of factual variants: by requiring a single program to succeed on most or all variants during rl training, we eliminate cases where an incorrect reasoning path nonetheless produces the right answer by accident, directly enforcing true consistency.  We then prompt the model to emit a compact program that issues only simple, atomic \texttt{retrieve} calls for each substep and uses a rejection filter to block any multi-step or invalid queries. This design uses the same simple \texttt{retrieve} calls for all cohort members, so the program runs unchanged on each variant; any change in output comes only from different facts, not from different reasoning steps. Finally, we apply Group Relative Policy Optimization (GRPO) \citep{shao2024deepseekmath} to maximize a composite reward—balancing cohort-level accuracy, retrieval-efficiency bonuses, and penalties for rejected lookups—across the entire cohort.  This pipeline compels the model to learn uniform, verifiable reasoning procedures rather than exploiting shortcuts on individual instances.

In experiments on five challenging reasoning benchmarks—ARC-Easy, ARC-Challenge \citep{clark2018arc}, StrategyQA \citep{geva2021strategyqa}, HotpotQA, and CommonsenseQA \citep{talmor-etal-2019-commonsenseqa}—\method{} consistently outperforms both vanilla and SFT baselines. We observe absolute improvements ranging from approximately 5\% to 10\% under lenient consistency criteria (requiring $\ge$4/6 similar questions correct) and 3\% to 8\% under strict criteria ($\ge$5/6 similar questions correct). Ablations confirm the importance of similar-question training, and human evaluation shows a 47\% win rate for our model’s reasoning paths, demonstrating that cohort-based RL with structured retrieval effectively enhances both accuracy and robustness across diverse reasoning tasks.

\textbf{Contributions.}  We summarize our key contributions:
\begin{itemize}
  \item \textbf{Cohort-Based RL Framework:} We introduce \method{}, a reinforcement learning approach that trains on cohorts of similar questions—each represented as an executable program—to explicitly enforce reasoning consistency across surface variants.
  \item \textbf{Structured Reward and Rejection Mechanism:} We design a composite reward that balances cohort-level accuracy, appropriate retrieval usage, and rejection penalties for trivial or multi-step queries, thereby discouraging shortcut behaviors.
  \item \textbf{End-to-End Data Pipeline:} We develop a high-quality pipeline for masked abstraction, automatic variant instantiation, and program synthesis, supported by cross-model and human verification.
  \item \textbf{Empirical Validation:} Comprehensive evaluations across five benchmarks show consistent improvement over strong baselines, with ablations and human studies further validating our approach.
\end{itemize}

\section{Related Work}

\paragraph{Reasoning Consistency}  
Large language models often exhibit inconsistent reasoning when faced with paraphrased or logically equivalent inputs. For example, prompts with similar surface familiarity but different underlying complexity yield divergent performance \citep{Li2024FamiCom}, and models may exploit spurious semantic cues rather than following intended chains of reasoning \citep{li2024deceptivesemanticshortcutsreasoning}. To mitigate this, researchers have added training regularizers or auxiliary losses for paraphrase‐invariance \citep{elazar2021consistency}, leveraged knowledge graphs to generate paired questions for fine‐tuning \citep{Rajan2024KonTest}, and applied self‐consistency decoding to vote out illogical paths \citep{wang2023selfconsistency}. Despite these efforts—and evidence that larger models do not automatically improve in logical consistency \citep{jang2022becel}—maintaining reasoning invariance across similar questions remains an open challenge. Our work enforces consistent, programmatic reasoning across groups of similar questions that share the same reasoning path.

\paragraph{Programmatic Abstractions for Reasoning}  
Programmatic or symbolic abstractions introduce a formal structure that can be executed and verified, improving transparency and reliability. \citep{zhou2024conceptualunbiasedreasoninglanguage, zhou2025selfsupervisedanalogicallearningusing} translate questions into symbolic programs for verifiable reasoning, while \citep{hong2024abstractionofthoughtmakeslanguagemodels} requires an Abstraction‐of‐Thought plan before refining a concrete solution. These approaches show that enforcing structured intermediate representations reduces reliance on brittle shortcuts and enhances problem-solving robustness.

\paragraph{Reinforcement Learning for Enhanced Reasoning.}
Reinforcement learning (RL) objectives can substantially boost an LLM’s ability to solve multi-step problems by optimizing the reasoning path. 
Verifiable intermediate rewards catch and correct logical mistakes, leading to more stable reasoning \citep{xu2025dro}.  
Composite reward functions that blend answer accuracy, factuality, and faithfulness yield more dependable outputs \citep{wang2024f2rl}.  
Potential‐based shaping adds domain priors as soft constraints, speeding up training while keeping policies optimal \citep{nguyen2020shaping}.  
Together, these advances in RL-driven reasoning path optimization, verifiable reward design, and structured shapinginspire our cohort-level consistency framework.

\section{Method}
\label{sec:method}

\begin{figure}[t]
  \centering
  \includegraphics[width=\linewidth]{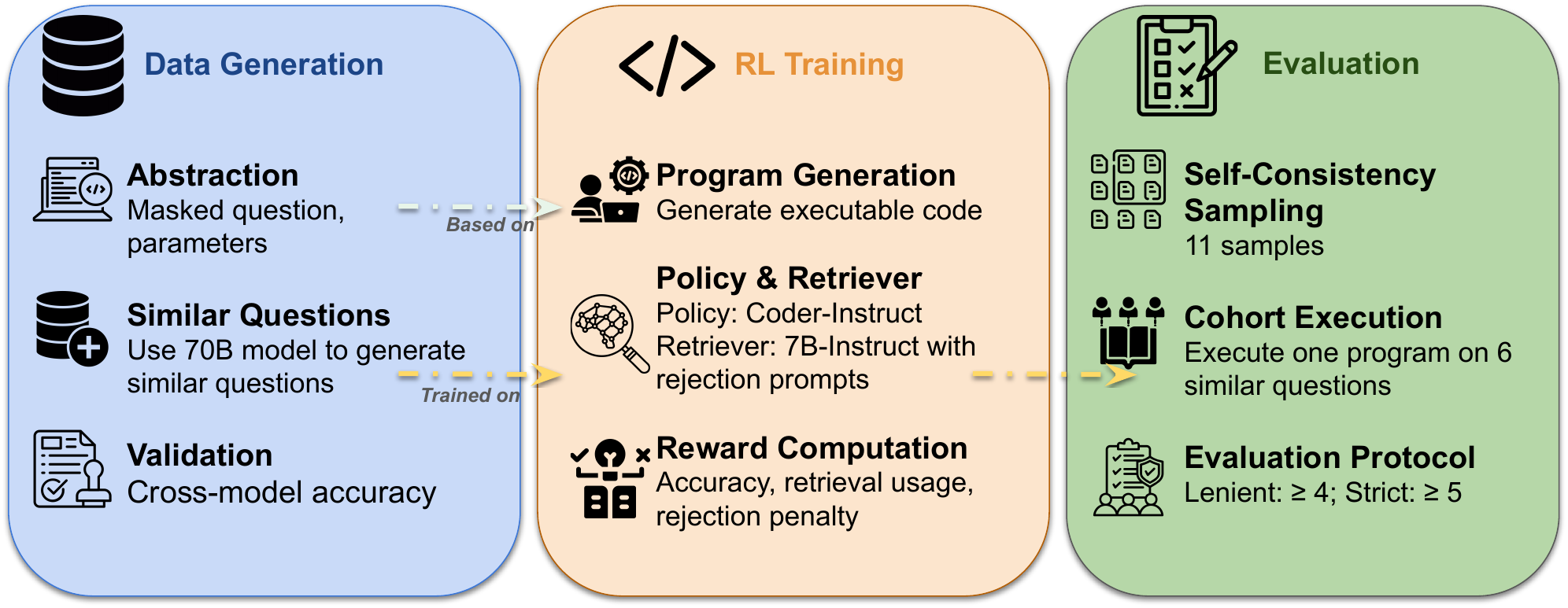}
  \caption{Overview of the CC-Learn pipeline. \textbf{Data Generation:} we abstract each question into a masked template with parameters, generate cohorts of similar variants using a 70B LLM, and validate answers via cross-model agreement. \textbf{RL Training:} we synthesize executable programs, use a Coder-Instruct policy and a 7B-Instruct retriever with rejection prompts, and compute a composite reward from accuracy, retrieval usage, and rejection penalties. \textbf{Evaluation:} we perform self-consistency sampling, execute each program on its cohort of six variants, and report performance under lenient (at least 4 of 6 correct) and strict (at least 5 of 6 correct) consistency criteria.}
  \label{fig:teaser}
\end{figure}

In this section, we introduce our framework for training language models to execute transparent reasoning through structured, executable programs. Our approach leverages a similar question generation strategy, rejection prompting, and a tailored RL scheme. This combination is designed to penalize shortcuts and reward consistent performance across similar questions, thereby compelling models to develop generalizable reasoning procedures rather than relying on superficial pattern-matching heuristics.

\subsection{Program Generation}
We represent each reasoning path as a small Python function of the form:
\begin{verbatim}
def answer(param1: Type1, …, paramK: TypeK) -> int:
    # sequence of retrieve(...) calls and control flow
\end{verbatim}
Only atomic \texttt{retrieve(question: str, type)} calls and control flows are allowed.  The model is given four inputs:
\begin{itemize}
  \item \emph{Masked question}: the template preserving logical structure (e.g.,  
    “Are \texttt{Film1} and \texttt{Film2} both \texttt{DocumentaryType} films that involve \texttt{Country}?”).
  \item \emph{Parameters}: the placeholder names (e.g., \texttt{Film1}, \texttt{Country}).
  \item \emph{Options}: the answer choices (e.g., \texttt{['No','Yes']}).
  \item \emph{Function header}: the exact \texttt{def answer(...)\ -> int} signature.
\end{itemize}
We guide the model’s output format with four few‐shot examples—covering boolean checks, numeric comparisons, list‐based loops, and dependent lookups—selected to span our cohort templates (see Appendix~\ref{appendix:rl_prompt}).  During generation, the policy model fills in the function body; at execution time, each \texttt{retrieve} call is issued to the retriever (with rejection filtering) to complete the lookup.

\subsection{Data Generation and Preprocessing}

To foster generalizable reasoning and provide a strong foundation for our models, we employ a high-quality data preparation pipeline.

\paragraph{Similar Questions Generation}
Central to our approach is the construction of cohorts of similar questions that share same reasoning paths but differ in factual content. This process begins with a corpus of 5,000 original questions from five domains (1,000 questions per domain). As \autoref{fig:example} shows, for each original question, we generate an abstraction by creating an abstraction template that preserves its core reasoning structure while parameterizing entities, allowing for the substitution of different facts(see Appendix ~\ref{appendix:abs_prompt}). Using LLaMA-3.3-70B-Instruct \citep{grattafiori2024llama3herdmodels}, we then instantiate 5 similar questions for each abstraction(see Appendix ~\ref{appendix:sim_prompt}). These variants are solvable through the same reasoning path dictated by the abstraction template but feature different specific entities and details. To ensure the fidelity of our training data, answers to all generated similar questions are cross-validated using three state-of-the-art models: LLaMA-3.3-70B-Instruct, 
DeepSeek-R1-Distill-Llama-70B\citep{deepseekai2025deepseekr1incentivizingreasoningcapability}, and Qwen-2.5-72B-Instruct\citep{qwen2, qwen2.5}. This process is supplemented by human verification of both the validity of similar questions and the correctness of their labels, as detailed in ~\ref{subsec:human_sim}.

\paragraph{Program Corpus for Supervised Fine-Tuning (SFT).}
\label{sec:sft_data}
To establish a robust baseline, we create a high-quality corpus of 500 exemplar programs (100 per domain), generated using LLaMA-3.3-70B-Instruct prompted with few-shot examples. Each program in this corpus is rigorously verified to ensure a 0\% rejection rate and 100\% execution accuracy on its corresponding question.

\subsection{Cohort-based Reinforcement Learning}

We train the policy model using Group Relative Policy Optimization (GRPO) on cohorts of similar questions.  Based on abstractions, the policy model generates an executable program, runs it on all variants (with a retriever handling simple lookups and rejecting complex queries), and receives a single, composite cohort-level reward.  This reward drives accuracy across the cohort, efficient retrieval usage, and discourages invalid lookups, guiding the model toward consistent, verifiable reasoning.

\paragraph{Model Architecture}

Our framework employs two complementary language models: \textbf{(a) Policy Model:} We utilize Qwen-2.5-7B-Coder-Instruct\citep{qwen2, qwen2.5, hui2024qwen2} as our primary model. It is tasked with generating structured, executable programs in response to the abstracted questions. This model is responsible for formulating the reasoning path. \textbf{(b) Retriever Model:} Qwen-2.5-7B-Instruct serves as our retriever model. It executes the simple retrieval calls generated within the programs by the policy model, specifically during the reward computation phase of GRPO. This architectural separation ensures that the policy model must formulate its reasoning strategy without direct access to factual information during program generation. 

\paragraph{Rejection Prompts in Retrieval}
To encourage the policy model to learn robust, generalizable reasoning and  prevent the policy model from circumventing the intended reasoning process by issuing trivial or multi‐step queries—behavior akin to the “deceptive shortcuts” observed in prior work \citep{li2024deceptivesemanticshortcutsreasoning}—we equip the retriever model with a rejection‐prompt filter. With a few‐shot prompt (see Appendix~\ref{appendix:rej_prompt}), the retriever only accepts straightforward, single‐step factual questions (e.g., Is \{Film1\} a \{DocumentaryType\} film?) and replies with “idk” to any multi‐step or invalid queries. Any rejected call will incur the penalty \(R_{\mathrm{rej}}\), thereby incentivizing the policy model to offload only elementary lookups and to internalize the full reasoning chain within the generated program.

\paragraph{Interactive Training Pipeline.}
The GRPO training loop proceeds interactively. First, the policy model proposes a program in response to an abstracted question. Second, this program is executed on both the original question and its five similar variants. Any information retrieval calls embedded in the program are handled by the retriever model. Finally, based on the execution outcomes across the entire question family, a scalar reward is computed. This reward considers accuracy, retrieval efficiency, and the rejection ratios.

\paragraph{Reward Components.}
Our composite reward function $R$ is designed to guide the model towards effective and generalizable reasoning. It is defined as $R = R_{\text{acc}} + R_{\text{ret}} + R_{\text{rej}}$, where the components are:
\begin{itemize}
    \item \textbf{Accuracy Reward ($R_{\text{acc}}$):} This reward is calculated as $R_{\text{acc}} = 0.2 \cdot n_{\text{correct}}$, where $n_{\text{correct}} \in \{0, 1, \dots, 6\}$ is the number of correctly answered questions within the cohort of similar questions (1 original + 5 similar). This yields $R_{\text{acc}} \in [0, 1.2]$.
    \item \textbf{Retrieve-Function Usage Reward ($R_{\text{ret}}$):} This component encourages appropriate problem decomposition and is assigned based on the number of retrieval calls ($n_{\text{calls}}$) made by the program:
    $$ R_{\text{ret}} = \begin{cases} -0.6 & \text{if } n_{\text{calls}} = 0 \\ \phantom{-}0 & \text{if } n_{\text{calls}} = 1 \\ +0.6 & \text{if } n_{\text{calls}} > 1 \end{cases} $$
    This results in $R_{\text{ret}} \in \{-0.6, 0, 0.6\}$. The retrieve-function usage reward encourages problem decomposition and discourages trivial solutions with no retrieval calls.
    \item \textbf{Rejection Penalty ($R_{\text{rej}}$):} This penalty discourages ineffective retrieval calls and is given by $R_{\text{rej}} = -0.1 \cdot n_{\text{rejected}}$, where $n_{\text{rejected}} \in \{0, 1, \dots, 6\}$ is the number of questions in the group whose retrieve call is rejected by the retrieve model. This leads to $R_{\text{rej}} \in [-0.6, 0]$. The Rejection Penalty penalizes attempts to re-ask the original question or formulate overly similar questions, forcing multi-step, valid reasoning.
\end{itemize}
The retrieve function reward is crucial for encouraging the model to break down complex problems into simpler, verifiable steps via appropriate information retrieval calls, rather than attempting to solve them monolithically or exploit loopholes (e.g., generating an empty program). The rejection mechanism explicitly blocks attempts to shortcut the reasoning process, such as re-asking the input question, thereby forcing genuine problem decomposition.

\paragraph{RL Variants}
We explore two variants of our RL approach to assess different aspects of learning:
\textbf{(a) Cohort Accuracy:} Accuracy-based rewards ($R_{\text{acc}}$) are granted only if the generated program successfully answers at least 4 out of the 6 questions in one group. This enforces a higher standard of generalizability. \textbf{(b) Normal Accuracy:} Accuracy rewards ($R_{\text{acc}}$) are granted for every successful program execution on a question within the group, regardless of performance on other questions in that group. This provides a more granular learning signal.


\subsection{Cohort Execution Test}
\label{subsec:method:eval_proto}

We evaluated each generated program on its cohort of six questions (the original plus five variants), and consistency measured the results under two criteria:
\begin{itemize}
    \item \textbf{Strict Accuracy:} A program is considered correct only if it successfully answers at least \textbf{5} out of the 6 questions in the group.
    \item \textbf{Lenient Accuracy:} A program is considered correct if it successfully answers at least \textbf{4} out of the 6 questions in the group.
\end{itemize}

\section{Experiments}
\label{sec:experiments}

In this section, we present comprehensive experimental details and results demonstrating the efficacy of the RL framework for training LLMs to perform transparent reasoning through structured, executable programs.

\subsection{Baselines}
We compare our Cohort-based Reinforcement Learning models against two baseline configurations to demonstrate the efficacy of our approach: \textbf{(a) Vanilla Model:} The off-the-shelf Qwen-2.5-Coder-Instruct checkpoint, used without any additional supervised fine-tuning or RL, serves as a strong pretrained LLM reference.
\textbf{(b) Supervised Fine-Tuning (SFT):} The Qwen-2.5-Coder-Instruct model further trained on our curated corpus of 500 high-quality exemplar programs (see \autoref{sec:sft_data}) until the held-out evaluation loss stabilized—typically around epoch 1.1—representing conventional program-level supervision.

\subsection{Hyperparameters for RL}
We train the RL model using the VERL framework \citep{Sheng_2025} with a learning rate of $1\times10^{-6}$, a KL-divergence loss coefficient of 0.001, 5 rollouts per update, and a training batch size of 128. The temperature for the rollouts is 1.

\subsection{Experimental Setup}

\paragraph{Datasets and Test Set}
We evaluate on five publicly available benchmarks: ARC‐Easy, ARC‐Challenge, CSQA, StrategyQA, and HotpotQA.  Our test set comprises 2,500 questions in total (500 per dataset), each paired with five similar questions generated by our abstraction pipeline.

\paragraph{Evaluation Protocol}
For each of the 2,500 test questions, we sample the policy model 11 times to assess self-consistency, execute each resulting program on its six-question cohort, and then compute both strict and lenient accuracy as defined in \autoref{subsec:method:eval_proto}. This protocol enables us to report not only standard accuracy but also the critical generalization-across-variants metrics.

\subsection{Main Results}
\label{subsec:main_results}

Table~\ref{tab:main_results_lenient} reports the lenient consistency results . Under this criterion, both RL variants yield substantial gains over the baselines. On ARC‐Challenge, the Cohort RL model achieves \textbf{29.8\%}, a +10.8 point absolute improvement over the Vanilla baseline (19.0\%) and +10.0 points over SFT (19.8\%). The Normal RL variant closely follows at 29.2\%, already a +9.4 point gain over SFT. For ARC‐Easy, Normal RL leads at \textbf{41.6\%}, outperforming SFT (33.4\%) by +8.2 points, while Cohort RL scores 39.6\% (+6.2 points). On CSQA, Cohort RL attains \textbf{37.2\%}, improving by +5.2 over SFT (32.0\%), and StrategyQA sees a +4.0 point boost (16.0\% vs.\ 12.0\%). HotpotQA remains challenging: SFT holds a narrow edge at 46.8\%, with Cohort and Normal RL at 45.8\% and 45.6\%, respectively. These results demonstrate that even under a relaxed consistency threshold, our RL framework markedly enhances both accuracy and stability across diverse reasoning tasks.

Table~\ref{tab:main_results_strict} shows strict consistency results. Here, the advantages of cohort training become even more pronounced. On ARC‐Challenge, Cohort RL reaches \textbf{22.0\%}, a +7.6 point gain over SFT (14.4\%) and +8.4 over Vanilla (13.6\%). ARC‐Easy improves to 30.0\% (+5.8 vs.\ SFT), CSQA to 28.8\% (+3.8), and StrategyQA to 7.8\% (+1.6). HotpotQA strict accuracy climbs to 29.0\%, +1.2 over SFT. Across all five benchmarks, the Cohort RL variant either matches or exceeds the Normal RL variant, indicating that enforcing a group‐level reward threshold helps the model learn more robust, generalizable reasoning procedures. In summary, our CC-Learn approach consistently outperforms both vanilla pretraining and supervised program tuning, achieving the highest accuracy and the greatest improvements in reasoning consistency under both evaluation criteria.

\begin{table}[t]
\centering
\setlength{\tabcolsep}{4pt}
\begin{tabular}{lccccc}
\toprule
Model    & ARC-Challenge         & ARC-Easy             & CSQA                 & StrategyQA           & HotpotQA             \\
\midrule
Vanilla  & $19.0 \pm 3.1$        & $30.0 \pm 4.0$       & $29.8 \pm 4.0$       & $12.6 \pm 2.9$       & $45.0 \pm 4.3$       \\
SFT      & $19.8 \pm 3.5$        & $33.4 \pm 4.1$       & $32.0 \pm 4.1$       & $12.0 \pm 2.9$       & $\mathbf{46.8 \pm 4.4}$ \\
Normal   & $\underline{29.2 \pm 4.0}$ & $\mathbf{41.6 \pm 4.3}$ & $\underline{36.0 \pm 4.2}$ & $\underline{14.8 \pm 3.1}$ & $45.6 \pm 4.4$       \\
Cohort     & $\mathbf{29.8 \pm 4.0}$    & $\underline{39.6 \pm 4.3}$ & $\mathbf{37.2 \pm 4.2}$    & $\mathbf{16.0 \pm 3.2}$    & $\underline{45.8 \pm 4.4}$ \\
\bottomrule
\end{tabular}
\caption{Lenient Accuracy (\%) across reasoning datasets using different training configurations: \textbf{Vanilla} is the off-the-shelf Qwen-2.5-Coder-Instruct model with no additional tuning; \textbf{SFT} is the same model supervised fine-tuned on our 500 exemplar programs; \textbf{Normal} is the RL variant receiving per-question accuracy rewards; and \textbf{Cohort} is our cohort-level RL model. \textbf{Bold} = best, \underline{underline} = second best.}
\label{tab:main_results_lenient}

\end{table}

\begin{table}[t]
\centering
\setlength{\tabcolsep}{4pt}
\begin{tabular}{lccccc}
\toprule
Model    & ARC-Challenge         & ARC-Easy             & CSQA                 & StrategyQA           & HotpotQA             \\
\midrule
Vanilla  & $13.6 \pm 2.9$        & $22.6 \pm 3.7$       & $20.6 \pm 3.5$       & $6.8 \pm 2.2$        & $27.2 \pm 3.9$       \\
SFT      & $14.4 \pm 3.1$        & $24.2 \pm 3.7$       & $25.0 \pm 3.8$       & $6.2 \pm 2.1$        & $27.8 \pm 3.9$       \\
Normal   & $\underline{20.8 \pm 3.6}$ & $\mathbf{30.6 \pm 4.0}$ & $\underline{28.4 \pm 4.0}$ & $\underline{7.0 \pm 2.3}$  & $\underline{28.0 \pm 3.9}$ \\
Cohort     & $\mathbf{22.0 \pm 3.6}$    & $\underline{30.0 \pm 4.0}$ & $\mathbf{28.8 \pm 4.0}$    & $\mathbf{7.8 \pm 2.3}$     & $\mathbf{29.0 \pm 4.0}$    \\
\bottomrule
\end{tabular}
\caption{Strict Accuracy (\%) across datasets. \textbf{Bold} = best, \underline{underline} = second best. We use the same terminology here as for lenient accuracy.}
\label{tab:main_results_strict}

\end{table}

\subsection{Sanity Check Experiments}
\paragraph{Reject Prompts Analysis}
\label{subsec:rejection_analysis}

To evaluate the validity of our rejection prompts for the retriever model in guiding the policy model's reasoning, we evaluated the rejection prompts on the 5000 training questions to get the rejection ratio of the original questions. We also evaluated the rejection prompts on simple questions - SimpleQA \citep{wei2024measuringshortformfactualitylarge} - to avoid over-rejection. Table~\ref{tab:rejection_rates} shows the rejection rates for multi-step complex questions compared to simple questions.

\begin{table}[h!]
\centering

\begin{tabular}{lc}
\toprule
Category & Rejection Rate(\%) \\
\midrule
ARC-CHALLENGE & 72.8 \\
ARC-EASY & 79.7 \\
CSQA & 70.1 \\
StrategyQA & 65.7 \\
HotpotQA & 71.8 \\
Overall (Multi-step Complex Questions) & 72.0 \\
\midrule
SimpleQA & 46.0 \\
\bottomrule
\end{tabular}
\caption{Rejection rates during RL training for different question categories.}
\label{tab:rejection_rates}

\end{table}

The significant differences in rejection rates between complex multi-step questions (72.0\%) and simple questions (46.0\%) confirm that our rejection mechanism effectively distinguishes between different question types. This behavior is crucial for preventing the model from circumventing the intended reasoning process by generating rephrasing retrieval calls, thereby forcing it to develop more sophisticated reasoning strategies.

\paragraph{Upper Bound Analysis}
To assess the theoretical feasibility of our evaluation protocol, we estimate a theoretical upper‐bound performance by running pass@128 on a random subset of 50 questions per domain and then manually correcting any incorrect programs. As shown in Table~\ref{tab:upper_bound}, after this minor intervention the model reaches near‐perfect accuracy—96.0\% on both ARC‐Challenge and ARC‐Easy, 88.0\% on CSQA, 92.0\% on StrategyQA, and 100.0\% on HotpotQA. Crucially, this demonstrates that the evaluation itself is not an insurmountable “mission impossible”; rather, the large gap between these upper‐bound scores and our current best results underscores that existing LLMs still fall well short of their potential and require significant advances in reasoning consistency and generalization.

\begin{table}[!htbp]
\centering

\begin{tabular}{lccc}
\toprule
Domain         & Accuracy & Correct Num & Total Num \\
\midrule
ARC-Challenge  & 96.0     & 48      & 50    \\
ARC-Easy       & 96.0     & 48      & 50    \\
CSQA           & 88.0     & 44      & 50    \\
StrategyQA     & 92.0     & 46      & 50    \\
HotpotQA       & 100.0     & 50      & 50    \\
\bottomrule
\end{tabular}
\caption{Estimated upper‐bound accuracy(\%) after pass@128 sampling and manual program correction.}
\label{tab:upper_bound}

\end{table}

\subsection{Ablation Studies}
\label{subsec:ablation_studies}

\paragraph{Training on Original Questions Only.}
To further evaluate the effectiveness of the similar question during training, we conducted an ablation study where RL training was performed solely on the original questions without considering the similar variants. \autoref{tab:g3_org_comparison_lenient} and \autoref{tab:g3_org_comparison_strict} compare the performance of this approach (Org) with our Cohort RL variant(Cohort).

\begin{table}[h!]
\centering
\setlength{\tabcolsep}{4pt}

\begin{tabular}{lccccc}
\toprule
Model & ARC‐Challenge & ARC‐Easy & CSQA & StrategyQA & HotpotQA \\
\midrule
Org & $25.4 \pm 3.8$        & $38.8 \pm 4.3$        & $37.0 \pm 4.2$        & $14.2 \pm 3.1$        & $\mathbf{45.8 \pm 4.4}$ \\
Cohort & $\mathbf{29.8 \pm 4.0}$ & $\mathbf{39.6 \pm 4.3}$ & $\mathbf{37.2 \pm 4.2}$ & $\mathbf{16.0 \pm 3.2}$ & $\mathbf{45.8 \pm 4.4}$ \\
\bottomrule
\end{tabular}
\caption{Lenient accuracy (\%) comparison between original (Org) and Cohort RL variant(Cohort). Bold = best}
\label{tab:g3_org_comparison_lenient}

\end{table}

\begin{table}[!htbp]
\centering
\setlength{\tabcolsep}{4pt}

\begin{tabular}{lccccc}
\toprule
Model & ARC‐Challenge & ARC‐Easy & CSQA & StrategyQA & HotpotQA \\
\midrule
Org & $19.4 \pm 3.5$        & $28.6 \pm 4.0$        & $28.2 \pm 4.0$        & $7.0 \pm 2.3$         & $27.6 \pm 3.9$        \\
Cohort & $\mathbf{22.0 \pm 3.6}$ & $\mathbf{30.0 \pm 4.0}$ & $\mathbf{28.8 \pm 4.0}$ & $\mathbf{7.8 \pm 2.3}$  & $\mathbf{29.0 \pm 4.0}$ \\
\bottomrule
\end{tabular}
\caption{Strict accuracy (\%) comparison between original (Org) and Cohort RL variant(Cohort). Bold = best.}
\label{tab:g3_org_comparison_strict}

\end{table}

The results demonstrate consistent performance degradation when training solely on original questions, particularly for complex reasoning tasks like ARC-Challenge (25.4\% vs. 29.8\% on lenient correctness). This confirms that the similar questions play a crucial role in compelling the model to learn generalizable reasoning strategies rather than question-specific shortcuts.

\paragraph{High-Quality Retriever}
We further investigate the impact of retrieval fidelity by replacing our default Qwen-2.5-7B-Instruct retriever with the more powerful Qwen3-32B-AWQ \citep{qwen3technicalreport} model during both training and evaluation. We evaluate three retrieval configurations: \textbf{7B–7B}, where both training and evaluation use the 7B retriever; \textbf{32B–7B}, where training employs the 32B retriever but evaluation uses the 7B retriever; and \textbf{32B–32B}, where both training and evaluation leverage the 32B retriever. As shown in \autoref{tab:high_quality_retriever_awq_lenient}, the 32B–32B configuration achieves the highest lenient accuracy across nearly all datasets, with particularly large gains on StrategyQA (34.8\% vs.\ 16.0\%) and HotpotQA (54.4\% vs.\ 45.8\%) compared to the 7B–7B baseline. Under strict consistency (\autoref{tab:high_quality_retriever_awq_strict}), the 32B–7B setup yields the best performance on ARC-Challenge (23.6\% vs.\ 22.0\%), while 32B–32B leads on ARC-Easy (31.0\% vs.\ 30.0\%), StrategyQA (15.4\% vs.\ 7.8\%), and HotpotQA (42.8\% vs.\ 29.0\%). These results underscore that higher-quality retrieval during both training and evaluation could further boost both accuracy and consistency, especially on complex, multi-hop reasoning tasks.

\begin{table}[!htbp]
  \centering
  \setlength{\tabcolsep}{4pt}

  \begin{tabular}{lccccc}
    \toprule
    Model        & ARC-Challenge & ARC-Easy & CSQA         & StrategyQA & HotpotQA \\
    \midrule
    7B-7B           & $29.8  \pm 4.0$        & $39.6  \pm 4.3$    & $\mathbf{37.2 \pm 4.2}$ & $16.0  \pm 3.2$     & $45.8 \pm 4.4$    \\
    32B-7B    & $31.6 \pm 3.8$         & $37.8 \pm 4.1$     & $35.8 \pm 3.9$       & $22.2 \pm 3.6$     & $48.2 \pm 3.7$     \\
    32B-32B      & $\mathbf{33.2 \pm 4.1}$ & $\mathbf{40.8 \pm 4.3}$ & $35.8 \pm 3.9$     & $\mathbf{34.8 \pm 4.2}$ & $\mathbf{54.4 \pm 4.4}$ \\
    \bottomrule
  \end{tabular}
  \caption{Lenient accuracy (\%) across reasoning datasets when using Qwen3-32B-AWQ as the retriever. Bold = best. Here, “7B-7B” denotes models trained and evaluated with the 7B retriever; “32B-7B” denotes models trained with the 32B retriever and evaluated with the 7B retriever; “32B-32B” denotes models trained and evaluated with the 32B retriever.}
  \label{tab:high_quality_retriever_awq_lenient}
  
\end{table}

\begin{table}[!htbp]
  \centering
  \setlength{\tabcolsep}{4pt}

  \begin{tabular}{lccccc}
    \toprule
    Model        & ARC-Challenge    & ARC-Easy     & CSQA           & StrategyQA    & HotpotQA     \\
    \midrule
    7B-7B           & $22.0 \pm 3.6$             & $30.0 \pm 4.0$         & $\mathbf{28.8 \pm 4.0}$  & $7.8  \pm 2.3$          & $29.0 \pm 4.0$         \\
    32B-7B    & $\mathbf{23.6 \pm 3.7}$    & $26.8 \pm 3.2$      & $26.6 \pm 3.5$           & $10.4 \pm 2.7$         & $34.4 \pm 3.9$        \\
    32B-32B       & $22.4 \pm 3.7$          & $\mathbf{31.0 \pm 4.0}$ & $27.8 \pm 3.9$         & $\mathbf{15.4 \pm 3.2}$ & $\mathbf{42.8 \pm 4.2}$\\
    \bottomrule
  \end{tabular}
  \caption{Strict accuracy (\%) across reasoning datasets when using Qwen3-32B-AWQ as the retriever. Bold = best. We use the same terminology here as for lenient accuracy.}
  \label{tab:high_quality_retriever_awq_strict}
  
\end{table}

\subsection{Human Study}

\paragraph{Similar Question Quality Assessment}

\label{subsec:human_sim}

To verify that our automated pipeline produces high-quality question variants, we assigned five annotators to assess a random sample of 150 generated questions (30 per domain) along two criteria: whether the answer label matches the ground truth, and whether the question follows its abstraction template. The detailed results are summarized in Table~\ref{tab:human_eval}.

\paragraph{Comparison of Reasoning Path}
We also conducted a comparison of programmatic reasoning paths from our RL‐trained model versus a supervised fine‐tuning (SFT) baseline. Two annotators evaluated 100 instances (20 per domain), choosing which program showed superior logical coherence, clarity, and decomposition. Results are included in Table~\ref{tab:human_eval}.

\begin{table}[!htbp]
\centering
\begin{tabular}{ccccc}
\toprule
\multicolumn{2}{c}{\itshape Accuracy (\%)} & \multicolumn{3}{c}{\itshape Win Rate (\%)} \\
\cmidrule(lr){1-2} \cmidrule(lr){3-5}
Answer Label    & Template Match & SFT     & Normal RL & Cohort RL \\
92.0            & 96.7           & 23.0    & 30.0      & 47.0      \\
\bottomrule
\end{tabular}
\caption{Human evaluation of generated questions and reasoning paths, reporting both accuracy metrics and win rates for different models.}
\label{tab:human_eval}
\end{table}

As shown in Table~\ref{tab:human_eval}, our pipeline achieves high label and template accuracy, and Cohort RL’s reasoning paths are preferred 47\% of the time over the left 2 models.

\section{Conclusion}

In this work, we introduced a novel training framework that substantially improves the consistency and reliability of language model reasoning. By leveraging cohorts of similar questions that share the same reasoning paths but differ in surface details, our approach explicitly teaches models to develop generalizable reasoning strategies rather than superficial shortcuts or brittle heuristics.

Our key innovation lies in the combination of executable program-based reasoning with a tailored RL scheme that rewards consistent performance across question variants. This approach addresses a fundamental challenge in LLM reasoning: the tendency to answer nearly identical questions inconsistently, which undermines reliability in real-world settings.

Experimental results across five diverse reasoning datasets demonstrate that our method significantly outperforms both vanilla and supervised fine-tuning baselines, with particularly pronounced improvements on complex multi-step reasoning tasks. The robust performance under strict correctness criteria (requiring correct answers to at least 5 out of 6 question variants) confirms that our approach successfully instills generalizable reasoning capabilities. Human evaluation further validates our approach, showing a preference (47\%) for the reasoning paths generated by our Cohort RL model over those from supervised fine-tuning. 

\section{Limitations}


\paragraph{Model Scale}  
Due to computational constraints, we did not utilize different policy models for RL training. We plan to investigate the applicability of our approach to larger or alternative models in future work.

\paragraph{Hyperparameter Tuning}
GRPO-based RL is resource‐heavy and slow on our hardware, so we used one heuristic hyperparameter setup instead of exhaustive tuning. This may understate our method’s true performance—future work should systematically tune these parameters to improve both accuracy and efficiency.

\newpage
\bibliographystyle{plain}
\bibliography{references}

\newpage
\appendix

\section{Prompts}

\subsection{Rejection Prompt}
\label{appendix:rej_prompt}
\begin{tcolorbox}[title=\footnotesize Prompt for \textit{Fact-Lookup Assistant},top=1mm,bottom=1mm]
\scriptsize
\textbf{System:} You are a fact-lookup assistant. For each user query, first decide if it's a simple, single-step fact lookup without solving it and then return a JSON object with exactly one key, "answer", wrapped in ```json ...```. Match the type specified in parentheses (int, str, list, bool). If a query requires more than a straightforward fact check or true/false lookup—for example, multi-step reasoning or subjective judgment—reply with "idk".

\medskip
\textbf{User:} Who finished immediately after the winner at the 1992 Olympic 100m final? (str)

\textbf{Assistant:} [Explanation] You must identify the winner, then determine who came second—this isn’t single-step.
```json
{"answer": "idk"}
```

\medskip
\textbf{User:} How many planets are in the solar system? (int)

\textbf{Assistant:} [Explanation] Simple fact check.
```json
{"answer": 8}
```

\medskip
\textbf{User:} What is the profession of Michael Jackson? (str)

\textbf{Assistant:} [Explanation] Single well-known profession of a public figure.
```json
{"answer": "singer"}
```

\medskip
\textbf{User:} Who has more than one Nobel Prize? (list)

\textbf{Assistant:} [Explanation] Factual list of individuals with multiple Nobel Prizes.
```json
{"answer": ["John Bardeen", "Frederick Sanger", "Linus Pauling", "Marie Curie"]}
```

\medskip
\textbf{User:} Is the CEO of Tesla older than the current President of France? (bool)

\textbf{Assistant:} [Explanation] Requires fetching and comparing two birthdates—multi-step.
```json
{"answer": "idk"}
```

\medskip
\textbf{User:} Is the Eiffel Tower located in Paris, France? (bool)

\textbf{Assistant:} [Explanation] Single-step landmark location.
```json
{"answer": false}
```

\medskip
\textbf{User:} Did England win any Olympic gold medals in 1800? (bool)

\textbf{Assistant:} [Explanation] Must check when the modern Olympics began and then medal records—multi-step.
```json
{"answer": "idk"}
```

\medskip
\textbf{User:} What is the population of the largest country entirely south of the equator? (int)

\textbf{Assistant:} [Explanation] Identify the country then lookup its population—multi-step.
```json
{"answer": "idk"}
```

\medskip
\textbf{User:} List the U.S. states admitted to the Union between the first and the last of the original 13 colonies. (list)

\textbf{Assistant:} [Explanation] Order states by admission date and filter—multi-step.
```json
{"answer": "idk"}
```

\medskip
\textbf{User:} Can food be cooked in the cosmic microwave background? (bool)

\textbf{Assistant:} [Explanation] Must compare CMB temperature (~2.7 K) to cooking physics—multi-step.
```json
{"answer": "idk"}
```

\medskip
\textbf{User:} Are Waris Hussein and Mathieu Kassovitz both actors? (bool)

\textbf{Assistant:} [Explanation] Fetch each person’s profession and compare—multi-step.
```json
{"answer": "idk"}
```

\medskip
\textbf{User:} \{prompt\} (\{return\_type\})

\textbf{Assistant:}
\end{tcolorbox}

\subsection{Abstraction Generation Prompt}
\label{appendix:abs_prompt}

\begin{tcolorbox}[title=\footnotesize Prompt for \textit{Masked Question Generation},top=1mm,bottom=1mm]
\scriptsize
\textbf{System:} Generate a masked question and corresponding parameters for the given question. These generations will be used to create similar questions that share the same reasoning path and options. The similar questions can only replace the values in the parameters. Therefore, be careful with the parameters. Do not include options or answers in the parameters. Present the output in JSON format.

\medskip
\textbf{User:} The Sugarcubes and SCH, are related to music? Options: A. No B. Yes. Answer: B

\textbf{Assistant:}
\begin{verbatim}
{
  "masked_question": "Are Entity1 and Entity2 both related to DomainX?",
  "parameters": {
    "Entity1": "The Sugarcubes",
    "Entity2": "SCH",
    "DomainX": "music"
  }
}
\end{verbatim}

\medskip
\textbf{User:} Are Ken Kwapis and Jan Švankmajer both filmmakers? Options: A. No B. Yes. Answer: B

\textbf{Assistant:}
\begin{verbatim}
{
  "masked_question": "Are Person1 and Person2 both ProfessionX?",
  "parameters": {
    "Person1": "Ken Kwapis",
    "Person2": "Jan Švankmajer",
    "ProfessionX": "filmmakers"
  }
}
\end{verbatim}

\medskip
\textbf{User:} Which of these is one similarity of the elements in the first group of the periodic table? Options: A. The elements in the group are very unreactive metals. B. The elements in the group react with oxygen in similar ways. Answer: B

\textbf{Assistant:}
\begin{verbatim}
{
  "masked_question": "Which of these is one similarity of the elements in the Group1 of the periodic table?",
  "parameters": {
    "Group1": "first group"
  }
}
\end{verbatim}

\medskip
\textbf{User:} Which celestial object listed below has the greatest density? Options: A. a planet B. a neutron star. Answer: B

\textbf{Assistant:}
\begin{verbatim}
{ 
  "masked_question": "Which celestial object listed below has the Superlative Property?", 
  "parameters": { 
    "Superlative": "greatest", 
    "Property": "density" 
  } 
}
\end{verbatim}

\medskip
\textbf{User:} On what would you serve an egg to someone? Options: A. plate B. chicken coop. Answer: A

\textbf{Assistant:}
\begin{verbatim}
{
  "masked_question": "On what would you serve a FoodItem to someone?",
  "parameters": {
    "FoodItem": "egg"
  }
}
\end{verbatim}

\medskip
\textbf{User:} Fighting inflation means being aware of many things, such as what with loans? Options: A. rising interest rates B. loss of money. Answer: A

\textbf{Assistant:}
\begin{verbatim}
{
  "masked_question": "Fighting EconomicIssue means being aware of many things, such as what with FinancialAspect?",
  "parameters": {
    "EconomicIssue": "inflation",
    "FinancialAspect": "loans"
  }
}
\end{verbatim}

\medskip
\textbf{User:} Question: \{question\} Options: \{options\}. Answer: \{answer\}

\textbf{Assistant:}

\end{tcolorbox}

\subsection{Similar Question Prompt}
\label{appendix:sim_prompt}
\begin{tcolorbox}[title=\footnotesize Prompt for \textit{Similar Question Generation},breakable,top=1mm,bottom=1mm]
\scriptsize
\textbf{System:} Generate five new questions that follow the same underlying reasoning process based on the masked questions, parameters, and the original question. Ensure that the choices and masked question remain the same as the original. You may only change the values within the \verb|parameters| dictionary. Present the output in JSON format.

\medskip
\textbf{User:} The Sugarcubes and SCH, are related to music? Options: A. No B. Yes. Answer: B.  
Masked Question: Are Entity1 and Entity2 both related to DomainX?  
Parameters: {'Entity1': 'The Sugarcubes', 'Entity2': 'SCH', 'DomainX': 'music'}.

\textbf{Assistant:}
\begin{verbatim}
{
  "similar_questions": [
    {
      "question": "Are The Beatles and ABBA both related to music?",
      "answer": "B",
      "parameters": {
        "Entity1": "The Beatles",
        "Entity2": "ABBA",
        "DomainX": "music"
      }
    },
    {
      "question": "Are Marvel and DC both related to comics?",
      "answer": "B",
      "parameters": {
        "Entity1": "Marvel",
        "Entity2": "DC",
        "DomainX": "comics"
      }
    },
    {
      "question": "Are NASA and SpaceX both related to space exploration?",
      "answer": "B",
      "parameters": {
        "Entity1": "NASA",
        "Entity2": "SpaceX",
        "DomainX": "space exploration"
      }
    },
    {
      "question": "Are Pepsi and Coca-Cola both related to beverages?",
      "answer": "B",
      "parameters": {
        "Entity1": "Pepsi",
        "Entity2": "Coca-Cola",
        "DomainX": "beverages"
      }
    },
    {
      "question": "Are Microsoft and Apple both related to technology?",
      "answer": "B",
      "parameters": {
        "Entity1": "Microsoft",
        "Entity2": "Apple",
        "DomainX": "technology"
      }
    }
  ]
}
\end{verbatim}

\medskip
\textbf{User:} Are Ken Kwapis and Jan Švankmajer both filmmakers? Options: A. No B. Yes. Answer: B.  
Masked Question: Are Person1 and Person2 both ProfessionX?  
Parameters: {'Person1': 'Christopher Nolan', 'Person2': 'Quentin Tarantino', 'ProfessionX': 'filmmakers'}.

\textbf{Assistant:}
\begin{verbatim}
{
  "similar_questions": [
    {
      "question": "Are Christopher Nolan and Quentin Tarantino both filmmakers?",
      "answer": "B",
      "parameters": {
        "Person1": "Christopher Nolan",
        "Person2": "Quentin Tarantino",
        "ProfessionX": "filmmakers"
      }
    },
    {
      "question": "Are J.K. Rowling and Stephen King both authors?",
      "answer": "B",
      "parameters": {
        "Person1": "J.K. Rowling",
        "Person2": "Stephen King",
        "ProfessionX": "authors"
      }
    },
    {
      "question": "Are Beyoncé and Elon Musk both singers?",
      "answer": "A",
      "parameters": {
        "Person1": "Beyoncé",
        "Person2": "Elon Musk",
        "ProfessionX": "singers"
      }
    },
    {
      "question": "Are Henry Ford and Albert Einstein both automobile entrepreneurs?",
      "answer": "A",
      "parameters": {
        "Person1": "Henry Ford",
        "Person2": "Albert Einstein",
        "ProfessionX": "automobile entrepreneurs"
      }
    },
    {
      "question": "Are Tony Hawk and Bob Burnquist both professional skateboarders?",
      "answer": "B",
      "parameters": {
        "Person1": "Tony Hawk",
        "Person2": "Bob Burnquist",
        "ProfessionX": "professional skateboarders"
      }
    }
  ]
}
\end{verbatim}

\medskip
\textbf{User:} Which of these is one similarity of the elements in the first group of the periodic table? Options: A. They all react similarly with water B. They have similar electron configurations. Answer: A.  
Masked Question: Which of these is one similarity of the elements in Group1 of the periodic table?  
Parameters: {'Group1': 'first group'}.

\textbf{Assistant:}
\begin{verbatim}
{
  "similar_questions": [
    {
      "question": "Which of these is one similarity of the elements in the 18th group of the periodic table?",
      "answer": "B",
      "parameters": {
        "Group1": "18th group"
      }
    },
    {
      "question": "Which of these is one similarity of the elements in the alkali metals of the periodic table?",
      "answer": "B",
      "parameters": {
        "Group1": "alkali metals"
      }
    },
    {
      "question": "Which of these is one similarity of the elements in the noble gases of the periodic table?",
      "answer": "B",
      "parameters": {
        "Group1": "noble gases"
      }
    },
    {
      "question": "Which of these is one similarity of the elements in the halogens of the periodic table?",
      "answer": "A",
      "parameters": {
        "Group1": "halogens"
      }
    },
    {
      "question": "Which of these is one similarity of the elements in the transition metals of the periodic table?",
      "answer": "B",
      "parameters": {
        "Group1": "transition metals"
      }
    }
  ]
}
\end{verbatim}

\medskip
\textbf{User:} Which celestial object listed below has the greatest density? Options: A. a planet B. a neutron star. Answer: B.  
Masked Question: Which celestial object listed below has the Superlative Property?  
Parameters: {'Superlative': 'greatest', 'Property': 'density'}.

\textbf{Assistant:}
\begin{verbatim}
{
  "similar_questions": [
    {
      "question": "Which celestial object listed below has the greatest density?",
      "answer": "A",
      "parameters": {
        "Superlative": "greatest",
        "Property": "density"
      }
    },
    {
      "question": "Which celestial object listed below has the highest temperature?",
      "answer": "A",
      "parameters": {
        "Superlative": "highest",
        "Property": "temperature"
      }
    },
    {
      "question": "Which celestial object listed below has the largest mass?",
      "answer": "B",
      "parameters": {
        "Superlative": "largest",
        "Property": "mass"
      }
    },
    {
      "question": "Which celestial object listed below has the brightest luminosity?",
      "answer": "B",
        "parameters": {
        "Superlative": "brightest",
        "Property": "luminosity"
      }
    },
    {
      "question": "Which celestial object listed below has the longest lifespan?",
      "answer": "B",
      "parameters": {
        "Superlative": "longest",
        "Property": "lifespan"
      }
    }
  ]
}
\end{verbatim}

\medskip
\textbf{User:} On what would you serve an egg to someone? Options: A. plate B. chicken coop. Answer: A.  
Masked Question: On what would you serve a FoodItem to someone?  
Parameters: {'FoodItem': 'egg'}.

\textbf{Assistant:}
\begin{verbatim}
{
  "similar_questions": [
    {
      "question": "On what would you serve an egg to someone?",
      "answer": "A",
      "parameters": {
        "FoodItem": "egg"
      }
    },
    {
      "question": "On what would you serve a soup to someone?",
      "answer": "B",
      "parameters": {
        "FoodItem": "soup"
      }
    },
    {
      "question": "On what would you serve a salad to someone?",
      "answer": "B",
      "parameters": {
        "FoodItem": "salad"
      }
    },
    {
      "question": "On what would you serve a sandwich to someone?",
      "answer": "A",
      "parameters": {
        "FoodItem": "sandwich"
      }
    },
    {
      "question": "On what would you serve a steak to someone?",
      "answer": "B",
      "parameters": {
        "FoodItem": "steak"
      }
    }
  ]
}
\end{verbatim}

\medskip
\textbf{User:} Fighting inflation means being aware of many things, such as what with loans? Options: A. rising interest rates B. loss of money. Answer: A.  
Masked Question: Fighting EconomicIssue means being aware of many things, such as what with FinancialAspect?  
Parameters: {'EconomicIssue': 'inflation', 'FinancialAspect': 'loans'}.

\textbf{Assistant:}
\begin{verbatim}
{
  "similar_questions": [
    {
      "question": "Fighting inflation means being aware of many things, such as what with loans?",
      "answer": "A",
      "parameters": {
        "EconomicIssue": "inflation",
        "FinancialAspect": "loans"
      }
    },
    {
      "question": "Fighting recession means being aware of many things, such as what with unemployment?",
      "answer": "B",
      "parameters": {
        "EconomicIssue": "recession",
        "FinancialAspect": "unemployment"
      }
    },
    {
      "question": "Fighting deflation means being aware of many things, such as what with prices?",
      "answer": "B",
      "parameters": {
        "EconomicIssue": "deflation",
        "FinancialAspect": "prices"
      }
    },
    {
      "question": "Fighting an economic crisis means being aware of many things, such as what with market volatility?",
      "answer": "A",
      "parameters": {
        "EconomicIssue": "economic crisis",
        "FinancialAspect": "market volatility"
      }
    },
    {
      "question": "Fighting a budget deficit means being aware of many things, such as what with government spending?",
      "answer": "B",
      "parameters": {
        "EconomicIssue": "budget deficit",
        "FinancialAspect": "government spending"
      }
    }
  ]
}
\end{verbatim}

\medskip
\textbf{User:} NOTE: Make sure the choices and masked question remain the same. Question: \{question\} Options: \{options\}. Answer: \{answer\}. Masked Question: \{masked\_question\}. Parameters: \{entities\}.

\textbf{Assistant:}
\end{tcolorbox}

\subsection{Program Generation Prompt For SFT}

\begin{tcolorbox}[title=\footnotesize Prompt for \textit{Answer Function Implementation},breakable,top=1mm,bottom=1mm]
\scriptsize
\textbf{System:} Implement the provided \texttt{answer(...)} function exactly as specified by the signature in Python. Return an integer corresponding to the index of the correct option in \texttt{Options}. Use the helper function \texttt{retrieve(question: str, answer\_type: Type) -> Any} only for simple fact-checking. Allowed \texttt{answer\_type} values: \texttt{int}, \texttt{float}, \texttt{list}, \texttt{str}, or \texttt{bool}. Do not embed reasoning logic inside \texttt{retrieve} calls—just simple fact-check questions. If a single \texttt{retrieve} question requires multi-step reasoning, decompose it into multiple simpler \texttt{retrieve} calls. Assume \texttt{retrieve} is implemented; do not define it. Include every input parameter in your \texttt{answer} function. Produce only the \texttt{answer} function—no extra code, comments, or imports beyond what's necessary.

\medskip
\textbf{User:} Masked Question: Are Entity1 and Entity2 both related to DomainX? Parameters: \texttt{dict\_keys(['Entity1', 'Entity2', 'DomainX'])}. Options: \texttt{['No', 'Yes']}. Function header: \texttt{def answer(Entity1: str, Entity2: str, DomainX: str) -> int}.

\textbf{Assistant:}
\begin{verbatim}
from typing import Any

def answer(Entity1: str, Entity2: str, DomainX: str) -> int:
    # First check Entity1's relationship
    entity1_related = retrieve(
        f"Is {Entity1} related to {DomainX}?", bool
    )
    # Then check Entity2's relationship
    entity2_related = retrieve(
        f"Is {Entity2} related to {DomainX}?", bool
    )
    # Both must be related for "Yes"
    if entity1_related and entity2_related:
        return 1
    else:
        return 0
\end{verbatim}

\medskip
\textbf{User:} Masked Question: Are both founders of CompanyX still involved in the company? Parameters: \texttt{dict\_keys(['CompanyX'])}. Options: \texttt{['No', 'Yes']}. Function header: \texttt{def answer(CompanyX: str) -> int}.

\textbf{Assistant:}
\begin{verbatim}
from typing import Any, List

def answer(CompanyX: str) -> int:
    founders = retrieve(
        f"Who are the founders of {CompanyX}?", list
    )
    still_involved = []
    for founder in founders:
        involved = retrieve(
            f"Is {founder} still involved with {CompanyX}?", bool
        )
        still_involved.append(involved)
    if all(still_involved):
        return 1
    else:
        return 0
\end{verbatim}

\medskip
\textbf{User:} Masked Question: Which of these is one similarity of the elements in Group1 of the periodic table? Parameters: \texttt{dict\_keys(['Group1'])}. Options: \texttt{['They all react similarly with water', 'They have similar electron configurations.']}. Function header: \texttt{def answer(Group1: str) -> int}.

\textbf{Assistant:}
\begin{verbatim}
from typing import Any, List

def answer(Group1: str) -> int:
    elements = retrieve(
        f"List the elements in {Group1} of the periodic table", list
    )
    valence_counts = []
    for el in elements:
        count = retrieve(
            f"How many valence electrons does {el} have?", int
        )
        valence_counts.append(count)
    if len(set(valence_counts)) == 1:
        return 1
    water_reactivities = []
    for el in elements:
        reactive = retrieve(
            f"Is {el} reactive with water?", bool
        )
        water_reactivities.append(reactive)
    if len(set(water_reactivities)) == 1:
        return 0
    return 0
\end{verbatim}

\medskip
\textbf{User:} Masked Question: Which celestial object listed below has the Superlative Property? Parameters: \texttt{dict\_keys(['Superlative', 'Property'])}. Options: \texttt{['a planet', 'a neutron star']}. Function header: \texttt{def answer(Superlative: str, Property: str) -> int}.

\textbf{Assistant:}
\begin{verbatim}
from typing import Any, List

def answer(Superlative: str, Property: str) -> int:
    planet_list = retrieve(
        f"Which known planet has the {Superlative} {Property}?", list
    )
    planet_name = planet_list[0]
    planet_value = retrieve(
        f"What is the {Property} of {planet_name} in standard units?", float
    )
    neutron_list = retrieve(
        f"Which known neutron star has the {Superlative} {Property}?", list
    )
    neutron_name = neutron_list[0]
    neutron_value = retrieve(
        f"What is the {Property} of {neutron_name} in the same units?", float
    )
    is_max = retrieve(
        f"Does '{Superlative}' mean selecting the highest {Property}?", bool
    )
    if is_max:
        return 0 if planet_value > neutron_value else 1
    is_min = retrieve(
        f"Does '{Superlative}' mean selecting the lowest {Property}?", bool
    )
    if is_min:
        return 0 if planet_value < neutron_value else 1
    return 0
\end{verbatim}

\medskip
\textbf{User:} Masked Question: On what would you serve a FoodItem to someone? Parameters: \texttt{dict\_keys(['FoodItem'])}. Options: \texttt{['plate', 'bowl']}. Function header: \texttt{def answer(FoodItem: str) -> int}.

\textbf{Assistant:}
\begin{verbatim}
from typing import Any

def answer(FoodItem: str) -> int:
    served_on_plate = retrieve(
        f"Is a plate a common item to serve {FoodItem}?", bool
    )
    served_on_bowl = retrieve(
        f"Is a bowl a common item to serve {FoodItem}?", bool
    )
    if served_on_plate:
        return 0
    else:
        return 1
\end{verbatim}

\medskip
\textbf{User:} Masked Question: Did HistoricalFigure have ethnicity closer to NativeGroup than ForeignGroup? Parameters: \texttt{dict\_keys(['HistoricalFigure', 'NativeGroup', 'ForeignGroup'])}. Options: \texttt{['No', 'Yes']}. Function header: \texttt{def answer(HistoricalFigure: str, NativeGroup: str, ForeignGroup: str) -> int}.

\textbf{Assistant:}
\begin{verbatim}
from typing import Any

def answer(HistoricalFigure: str, NativeGroup: str, ForeignGroup: str) -> int:
    has_native = retrieve(
        f"Did {HistoricalFigure} have ancestry from {NativeGroup}?", bool
    )
    has_foreign = retrieve(
        f"Did {HistoricalFigure} have ancestry from {ForeignGroup}?", bool
    )
    if has_native and not has_foreign:
        return 1
    if has_foreign and not has_native:
        return 0
    more_native = retrieve(
        f"Did {HistoricalFigure} have a greater proportion of {NativeGroup} ancestry than {ForeignGroup}?", bool
    )
    if more_native:
        return 1
    else:
        return 0
\end{verbatim}

\medskip
\textbf{User:} Masked Question: Which reaction did Person have after TimePeriod in a ColdEnvironment? Parameters: \texttt{dict\_keys(['Person','TimePeriod','ColdEnvironment','PhysicalReaction'])}. Options: \texttt{['chills','calm down']}. Function header: \texttt{def answer(Person: str, TimePeriod: str, ColdEnvironment: str, PhysicalReaction: str) -> int}.

\textbf{Assistant:}
\begin{verbatim}
from typing import Any

def answer(Person: str, TimePeriod: str, ColdEnvironment: str, PhysicalReaction: str) -> int:
    had_chills = retrieve(
        f"Did {Person} have chills after {TimePeriod} in a {ColdEnvironment}?", bool
    )
    if had_chills:
        return 0
    had_calm = retrieve(
        f"Did {Person} calm down after {TimePeriod} in a {ColdEnvironment}?", bool
    )
    if had_calm:
        return 1
    return 0
\end{verbatim}

\medskip
\textbf{User:} Masked Question: Which is a way that Earth loses energy from the EnergySource? Parameters: \texttt{dict\_keys(['EnergySource'])}. Options: \texttt{['reflection','absorption']}. Function header: \texttt{def answer(EnergySource: str) -> int}.

\textbf{Assistant:}
\begin{verbatim}
from typing import Any

def answer(EnergySource: str) -> int:
    source_desc = retrieve(
        f"How does Earth lose energy from {EnergySource}?", str
    )
    is_reflection = retrieve(
        f"Does '{source_desc}' indicate reflection?", bool
    )
    if is_reflection:
        return 0
    is_absorption = retrieve(
        f"Does '{source_desc}' indicate absorption?", bool
    )
    if is_absorption:
        return 1
    return 0
\end{verbatim}

\medskip
\textbf{User:} NOTE: You must use all input parameters, decompose complex queries into multiple \texttt{retrieve} calls, and ensure at least two \texttt{retrieve} calls per function. \{input\}

\textbf{Assistant:}
\end{tcolorbox}

\subsection{Program Generation Prompt For RL}
\label{appendix:rl_prompt}

\begin{tcolorbox}[title=\footnotesize Prompt for \textit{Answer Function Implementation},breakable,top=1mm,bottom=1mm]
\scriptsize
\textbf{System:} Implement the provided \texttt{answer(...)} function exactly as specified by the signature in Python. Return an integer corresponding to the index of the correct option in \texttt{Options}. Use the helper function \texttt{retrieve(question: str, answer\_type: Type) -> Any} only for simple fact-checking. Allowed \texttt{answer\_type} values: \texttt{int}, \texttt{float}, \texttt{list}, \texttt{str}, or \texttt{bool}. Do not embed reasoning logic inside \texttt{retrieve} calls—just simple fact-check questions. If a single \texttt{retrieve} question requires multi-step reasoning, decompose it into multiple simpler \texttt{retrieve} calls. Assume \texttt{retrieve} is implemented; do not define it. Include every input parameter in your \texttt{answer} function. Produce only the \texttt{answer} function—no extra code, comments, or imports beyond what's necessary.

\medskip
\textbf{User:} Masked Question: Are both founders of CompanyX still involved in the company? Parameters: \texttt{dict\_keys(['CompanyX'])}. Options: \texttt{['No', 'Yes']}. Function header: \texttt{def answer(CompanyX: str) -> int}.

\textbf{Assistant:}
\begin{verbatim}
from typing import Any, List

def answer(CompanyX: str) -> int:
    founders = retrieve(
        f"Who are the founders of {CompanyX}?", list
    )
    still_involved = []
    for founder in founders:
        involved = retrieve(
            f"Is {founder} still involved with {CompanyX}?", bool
        )
        still_involved.append(involved)
    if all(still_involved):
        return 1
    else:
        return 0
\end{verbatim}

\medskip
\textbf{User:} Masked Question: Which celestial object listed below has the Superlative Property? Parameters: \texttt{dict\_keys(['Superlative', 'Property'])}. Options: \texttt{['a planet', 'a neutron star']}. Function header: \texttt{def answer(Superlative: str, Property: str) -> int}.

\textbf{Assistant:}
\begin{verbatim}
from typing import Any, List

def answer(Superlative: str, Property: str) -> int:
    planet_list = retrieve(
        f"Which known planet has the {Superlative} {Property}?", list
    )
    planet_name = planet_list[0]
    planet_value = retrieve(
        f"What is the {Property} of {planet_name} in standard units?", float
    )
    neutron_list = retrieve(
        f"Which known neutron star has the {Superlative} {Property}?", list
    )
    neutron_name = neutron_list[0]
    neutron_value = retrieve(
        f"What is the {Property} of {neutron_name} in the same units?", float
    )
    is_max = retrieve(
        f"Does '{Superlative}' mean selecting the highest {Property}?", bool
    )
    if is_max:
        return 0 if planet_value > neutron_value else 1
    is_min = retrieve(
        f"Does '{Superlative}' mean selecting the lowest {Property}?", bool
    )
    if is_min:
        return 0 if planet_value < neutron_value else 1
    return 0
\end{verbatim}

\medskip
\textbf{User:} Masked Question: On what would you serve a FoodItem to someone? Parameters: \texttt{dict\_keys(['FoodItem'])}. Options: \texttt{['plate', 'bowl']}. Function header: \texttt{def answer(FoodItem: str) -> int}.

\textbf{Assistant:}
\begin{verbatim}
from typing import Any

def answer(FoodItem: str) -> int:
    served_on_plate = retrieve(
        f"Is a plate a common item to serve {FoodItem}?", bool
    )
    served_on_bowl = retrieve(
        f"Is a bowl a common item to serve {FoodItem}?", bool
    )
    if served_on_plate:
        return 0
    else:
        return 1
\end{verbatim}

\medskip
\textbf{User:} Masked Question: Did HistoricalFigure have ethnicity closer to NativeGroup than ForeignGroup? Parameters: \texttt{dict\_keys(['HistoricalFigure', 'NativeGroup', 'ForeignGroup'])}. Options: \texttt{['No', 'Yes']}. Function header: \texttt{def answer(HistoricalFigure: str, NativeGroup: str, ForeignGroup: str) -> int}.

\textbf{Assistant:}
\begin{verbatim}
from typing import Any

def answer(HistoricalFigure: str, NativeGroup: str, ForeignGroup: str) -> int:
    has_native = retrieve(
        f"Did {HistoricalFigure} have ancestry from {NativeGroup}?", bool
    )
    has_foreign = retrieve(
        f"Did {HistoricalFigure} have ancestry from {ForeignGroup}?", bool
    )
    if has_native and not has_foreign:
        return 1
    if has_foreign and not has_native:
        return 0
    more_native = retrieve(
        f"Did {HistoricalFigure} have a greater proportion of {NativeGroup} ancestry than {ForeignGroup}?", bool
    )
    if more_native:
        return 1
    else:
        return 0
\end{verbatim}

\medskip
\textbf{User:} NOTE: You must use all input parameters, decompose complex queries into multiple \texttt{retrieve} calls, and ensure at least two \texttt{retrieve} calls per function. \{\texttt{input}\}

\end{tcolorbox}

\section{RL training}

We train RL models on 2 H200s with VERL framework. For the retriever model, we use temperature=0.7. For the exact memory usage, we serve an Instruct server and also do the training on the same 2 H200 gpus. While doing the back propagation, the memory will reach full capacity. In inference time, it reaches half of the capacity.

\section{Example}
\label{sec:screenshot}
The screenshot for the example.

\begin{figure}[!htbp]
  \centering
  \includegraphics[width=\linewidth]{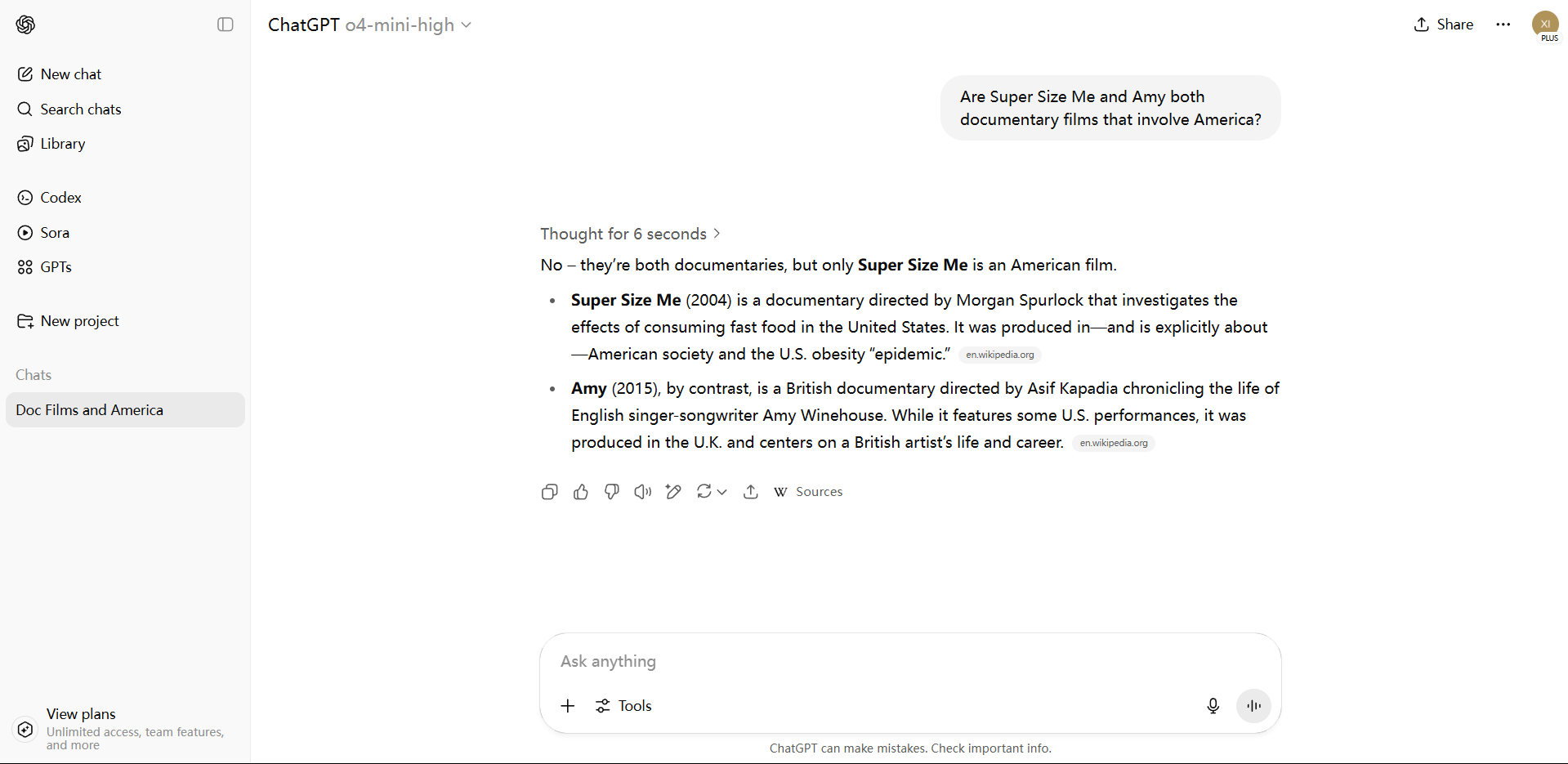}
  \caption{The screenshot for the example.}
  \label{fig:screenshot}
\end{figure}


\newpage
\section*{NeurIPS Paper Checklist}

The checklist is designed to encourage best practices for responsible machine learning research, addressing issues of reproducibility, transparency, research ethics, and societal impact. Do not remove the checklist: {\bf The papers not including the checklist will be desk rejected.} The checklist should follow the references and follow the (optional) supplemental material.  The checklist does NOT count towards the page
limit. 

Please read the checklist guidelines carefully for information on how to answer these questions. For each question in the checklist:
\begin{itemize}
    \item You should answer \answerYes{}, \answerNo{}, or \answerNA{}.
    \item \answerNA{} means either that the question is Not Applicable for that particular paper or the relevant information is Not Available.
    \item Please provide a short (1–2 sentence) justification right after your answer (even for NA). 
\end{itemize}

{\bf The checklist answers are an integral part of your paper submission.} They are visible to the reviewers, area chairs, senior area chairs, and ethics reviewers. You will be asked to also include it (after eventual revisions) with the final version of your paper, and its final version will be published with the paper.

The reviewers of your paper will be asked to use the checklist as one of the factors in their evaluation. While "\answerYes{}" is generally preferable to "\answerNo{}", it is perfectly acceptable to answer "\answerNo{}" provided a proper justification is given (e.g., "error bars are not reported because it would be too computationally expensive" or "we were unable to find the license for the dataset we used"). In general, answering "\answerNo{}" or "\answerNA{}" is not grounds for rejection. While the questions are phrased in a binary way, we acknowledge that the true answer is often more nuanced, so please just use your best judgment and write a justification to elaborate. All supporting evidence can appear either in the main paper or the supplemental material, provided in appendix. If you answer \answerYes{} to a question, in the justification please point to the section(s) where related material for the question can be found.


\begin{enumerate}

\item {\bf Claims}
    \item[] Question: Do the main claims made in the abstract and introduction accurately reflect the paper's contributions and scope?
    \item[] Answer: \answerYes{} 
    \item[] Justification: The main claim of the paper clearly reflects the contribution and applicaiton scope.

\item {\bf Limitations}
    \item[] Question: Does the paper discuss the limitations of the work performed by the authors?
    \item[] Answer: \answerYes{} 
    \item[] Justification: The Limitation section is the last section of this paper.

\item {\bf Theory assumptions and proofs}
    \item[] Question: For each theoretical result, does the paper provide the full set of assumptions and a complete (and correct) proof?
    \item[] Answer: \answerYes{} 
    \item[] Justification: The main claim of the paper, which is training a better reasoning path via our approach, is proved by the empirical results.

    \item {\bf Experimental result reproducibility}
    \item[] Question: Does the paper fully disclose all the information needed to reproduce the main experimental results of the paper to the extent that it affects the main claims and/or conclusions of the paper (regardless of whether the code and data are provided or not)?
    \item[] Answer: \answerYes{} 
    \item[] Justification: We will release the code on GitHub to facilitate reproducibility.

\item {\bf Open access to data and code}
    \item[] Question: Does the paper provide open access to the data and code, with sufficient instructions to faithfully reproduce the main experimental results, as described in supplemental material?
    \item[] Answer: \answerYes{} 
    \item[] Justification: We will release the code on GitHub to facilitate reproducibility. 

\item {\bf Experimental setting/details}
    \item[] Question: Does the paper specify all the training and test details (e.g., data splits, hyperparameters, how they were chosen, type of optimizer, etc.) necessary to understand the results?
    \item[] Answer: \answerYes{} 
    \item[] Justification: The training details, including prompts, hyperparameters are included in the appendix.

\item {\bf Experiment statistical significance}
    \item[] Question: Does the paper report error bars suitably and correctly defined or other appropriate information about the statistical significance of the experiments?
    \item[] Answer: \answerYes{} 
    \item[] Justification: Every table has a variance calculation.

\item {\bf Experiments compute resources}
    \item[] Question: For each experiment, does the paper provide sufficient information on the computer resources (type of compute workers, memory, time of execution) needed to reproduce the experiments?
    \item[] Answer: \answerYes{} 
    \item[] Justification: We included software specifications in the method section and hardware in the appendix section.
    
\item {\bf Code of ethics}
    \item[] Question: Does the research conducted in the paper conform, in every respect, with the NeurIPS Code of Ethics \url{https://neurips.cc/public/EthicsGuidelines}?
    \item[] Answer: \answerYes{} 
    \item[] Justification: We have read the Code of Ethics and made sure the paper conform to it.

\item {\bf Broader impacts}
    \item[] Question: Does the paper discuss both potential positive societal impacts and negative societal impacts of the work performed?
    \item[] Answer: \answerNA{} 
    \item[] Justification: The main contribution of this paper is a better training scheme for LLM. No negative social impacts.
    
\item {\bf Safeguards}
    \item[] Question: Does the paper describe safeguards that have been put in place for responsible release of data or models that have a high risk for misuse (e.g., pretrained language models, image generators, or scraped datasets)?
    \item[] Answer: \answerNA{} 
    \item[] Justification: There is no such risk for this paper.

\item {\bf Licenses for existing assets}
    \item[] Question: Are the creators or original owners of assets (e.g., code, data, models), used in the paper, properly credited and are the license and terms of use explicitly mentioned and properly respected?
    \item[] Answer: \answerNA{} 
    \item[] Justification: This paper does not use existing assets

\item {\bf New assets}
    \item[] Question: Are new assets introduced in the paper well documented and is the documentation provided alongside the assets?
    \item[] Answer: \answerNA{} 
    \item[] Justification: This paper does not release new assets.

\item {\bf Crowdsourcing and research with human subjects}
    \item[] Question: For crowdsourcing experiments and research with human subjects, does the paper include the full text of instructions given to participants and screenshots, if applicable, as well as details about compensation (if any)? 
    \item[] Answer: \answerNA{} 
    \item[] Justification: All annotations were performed internally by the paper’s authors; no external crowd‑workers or study participants were involved.

\item {\bf Institutional review board (IRB) approvals or equivalent for research with human subjects}
    \item[] Question: Does the paper describe potential risks incurred by study participants, whether such risks were disclosed to the subjects, and whether Institutional Review Board (IRB) approvals (or an equivalent approval/review based on the requirements of your country or institution) were obtained?
    \item[] Answer: \answerNA{} 
    \item[] Justification: The work does not meet the federal definition of human‑subjects research; therefore no IRB review was required.

\item {\bf Declaration of LLM usage}
    \item[] Question: Does the paper describe the usage of LLMs if it is an important, original, or non-standard component of the core methods in this research? Note that if the LLM is used only for writing, editing, or formatting purposes and does not impact the core methodology, scientific rigorousness, or originality of the research, declaration is not required.
    \item[] Answer: \answerYes{} 
    \item[] Justification: This paper is about a better training scheme for LLMs.

\end{enumerate}

\end{document}